\documentclass[11pt]{article}

\PassOptionsToPackage{table}{xcolor}

\usepackage[preprint]{acl}

\usepackage{times}
\usepackage{latexsym}

\usepackage[T1]{fontenc}
\usepackage[utf8]{inputenc}

\usepackage{microtype}

\usepackage{inconsolata}

\usepackage{graphicx}
\usepackage{kotex}
\usepackage{amsmath}
\usepackage{amssymb}
\usepackage{booktabs}
\usepackage{multirow}
\usepackage{subcaption}
\usepackage{xcolor}
\usepackage{newtxtext,newtxmath}
\usepackage{xcolor}
\usepackage{mdframed}
\usepackage{kotex}
\mdfsetup{%
linecolor=white,
backgroundcolor=gray!20, 
}
\usepackage{authblk} 
\usepackage{hyperref} 
\usepackage{calligra}
\usepackage[T1]{fontenc}
\usepackage[most]{tcolorbox}
\newtcolorbox{myrqbox}[1]{
  breakable,
  colback=white, % 배경색 (아주 연한 파랑)
  colframe=blue!5!white, % 테두리 색 (진한 파랑)
  fonttitle=\bfseries, % 제목 폰트 (굵게)
  coltitle=black, % 제목 색상 (검정)
  title=#1, % 박스 제목
  arc=2mm, % 모서리 둥글기
  boxsep=2pt, % 텍스트와 테두리 사이 간격
  left=6pt, % 왼쪽 여백
  right=6pt, % 오른쪽 여백
  top=4pt, % 위쪽 여백
  bottom=4pt % 아래쪽 여백
}

\usepackage[table]{xcolor}
\definecolor{lightgray}{gray}{0.9}
\usepackage{tcolorbox}
\tcbuselibrary{breakable}

\title{DLM-SWAI: Steering Diffusion Language Models Before They Unmask}

\author{
  \textbf{Hyeseon An,}
  \textbf{Yo-Sub Han} \thanks{Corresponding Author.} \\
  Department of Computer Science, Yonsei University, Seoul, South Korea \\
  \texttt{\{\href{mailto:hsan@yonsei.ac.kr}{hsan},
  \href{mailto:emmous@yonsei.ac.kr}{emmous}\}@yonsei.ac.kr}
}

\begin{document}
\maketitle

\begin{abstract}
Steering language model generation toward desired textual properties is essential for practical deployment, and inference-time methods are particularly appealing because they enable controllable generation without retraining. Recent work has also highlighted diffusion language models as an emerging generation paradigm with distinct decoding properties. However, most existing steering approaches either rely on auxiliary models or are designed for autoregressive next-token decoding, making them difficult to apply to diffusion language models~(DLMs), which generate text through iterative denoising of partially masked sequences. Therefore, we propose DLM-SWAI, a simple training-free steering method that biases the token distribution at each denoising step using pre-computed token-level style scores. Experiments on style and safety control tasks show that DLM-SWAI effectively steers diffusion language models while preserving generation quality and requiring minimal computational overhead. Ablations further reveal a controllable trade-off between steering strength and fluency, and our analysis links class-wise steerability to the strength of token-level attribute cues. Our code is available at \url{https://github.com/hsannn/dlm-swai}.
\end{abstract}

\section{Introduction}
% controllable generation이 왜 중요한가
Language models are increasingly expected not only to generate fluent and
relevant text, but also to do so in ways that satisfy user intent and
application-level constraints \citep{ouyang2022training}. In practical deployment,
desirable outputs often
depend on controllable properties such as style, tone, politeness, sentiment, or
safety \citep{bai2022constitutional}. As a result, controllable generation has become a core capability for
modern language models rather than a secondary feature. A model that merely
produces plausible text is often insufficient for real-world use; practical
systems must also support mechanisms for steering generation toward desired
textual attributes.

% inference-time steering이 왜 practical한가
Among the various approaches to controllable generation, inference-time steering
is particularly attractive because it enables flexible behavior control without
retraining the base model \citep{pan2025survey}. Such methods can be applied post hoc, reused across
multiple target attributes, and often incur substantially lower cost than
fine-tuning or training attribute-specific control modules. This makes
inference-time steering a practical solution for adapting a single base model to
diverse deployment settings.

\begin{table}[t]
\centering
\resizebox{\linewidth}{!}{%
\begin{tabular}{lccc}
\toprule
Method & \shortstack{Training-\\free} & \shortstack{No aux.\\model} & \shortstack{No hidden\\state} \\
\midrule
Classifier guidance & \checkmark & $\times$ & \checkmark \\
ILRR          & \checkmark & $\times$ & \checkmark \\
Activation steering& \checkmark & \checkmark & $\times$ \\
\textbf{DLM-SWAI (Ours)}                           & \checkmark & \checkmark & \checkmark \\
\bottomrule
\end{tabular}}
\caption{Design-space positioning of DLM-SWAI among representative DLM steering methods.\protect\footnotemark
Checkmarks indicate whether each method avoids training, auxiliary guidance models, or hidden-state access.}
\label{tab:positioning}
\end{table}
\footnotetext{Rows summarize representative prior methods: classifier guidance for discrete diffusion models \citep{nisonoff2025unlocking}, ILRR \citep{avrahami2026ilrr}, and activation steering for masked diffusion language models \citep{shnaidman2025activation}.}

% 그런데 기존 steering은 AR 중심이다
However, most existing steering methods have been developed primarily for
autoregressive language models. In many cases, they either assume left-to-right
next-token decoding or rely on auxiliary components such as classifiers, reward
models, or attribute predictors to provide guidance signals. These assumptions
make such approaches less suitable for diffusion language models (DLMs), which
generate text through iterative denoising of partially masked sequences rather
than sequential next-token prediction. As a result, methods designed around
autoregressive decoding do not naturally transfer to the diffusion setting,
while auxiliary-model-based approaches reduce the simplicity and efficiency that
make inference-time steering appealing in the first place. Table~\ref{tab:positioning}
summarizes this design-space gap for representative DLM steering methods.

% 그래서 왜 DLM에서도 따로 steering을 봐야 하는가
This limitation is increasingly important because diffusion language models are
emerging as a viable alternative generation paradigm with distinct decoding
behavior and promising generation capabilities \citep{gong2022diffuseq}.
The question is not whether DLMs
should replace autoregressive models, but whether controllability should be
available whenever DLMs are used as practical text generators \citep{arriola2025block}.
If controllable
generation is a deployment-critical capability, then DLMs must also support
lightweight and effective steering mechanisms. Yet despite the growing interest
in diffusion-based text generation, simple training-free methods for steering
DLM outputs remain underexplored.

% DLM 구조는 오히려 step-wise control에 잘 맞는다
At the same time, the iterative structure of diffusion decoding offers a natural
opportunity for control. Because a DLM repeatedly refines token distributions
over multiple denoising steps, guidance can be injected not only at individual
token choices but throughout the denoising trajectory. A token-level bias can
therefore affect both which tokens are sampled and which masked positions become
confident enough to be committed early. This suggests that effective steering in
DLMs may not require heavyweight auxiliary models or additional training, but
can instead be achieved by biasing the model's token predictions throughout the
denoising process.

% 그래서 우리는 DLM-SWAI를 제안한다
In this work, we propose \textbf{DLM-SWAI}, a \textbf{S}tatistical \textbf{W}riting style \textbf{A}ligned \textbf{I}nference for
\textbf{D}iffusion \textbf{L}anguage \textbf{M}odels.
DLM-SWAI uses pre-computed token-level style scores to bias the token
distribution predicted at each denoising step, enabling controllable generation
without auxiliary guidance models or parameter updates.
Although token-level logit biasing has been explored for autoregressive
decoding~\citep{an2026steering}, the way these scores act in a DLM differs
structurally: a single bias is injected into all masked positions at once and
shapes the unmasking order itself, so token-level preferences compound along the
denoising trajectory (Section~\ref{subsec:dlm_design}).
The method is lightweight, training-free, and naturally compatible with the
iterative denoising process of diffusion generation. Through experiments on
multiple style and safety control tasks, we show that DLM-SWAI effectively
steers DLM outputs toward desired attributes while preserving generation quality
and incurring only minimal computational overhead.

Our contributions are as follows:
\begin{itemize}
    \item We identify the lack of simple and diffusion-compatible inference-time
    steering methods as an important gap in controllable generation for
    diffusion language models.
    \item We propose DLM-SWAI, a training-free steering method that injects
    token-level control signals directly into the denoising process, and we
    characterize the two diffusion-specific mechanisms that distinguish it from
    autoregressive logit shifting.
    \item We demonstrate across style and safety control tasks that the proposed
    method provides effective controllability with low overhead while
    maintaining generation quality, verified with both fluency (perplexity) and
    semantic (BERTScore) metrics.
    \item We provide ablations and analyses that explain when and why the method
    works, quantifying the trade-off between steering strength and fluency and
    linking class-wise steerability to the distribution of token-level attribute
    cues.
\end{itemize}

\section{Related Work}
Model steering is important because a strong generative
model should not merely produce plausible text, but should also allow its
behavior to be directed toward desired properties. Without such controllability,
even capable language models remain difficult to deploy in settings where
outputs must satisfy task-specific, stylistic, or safety-related requirements.
We therefore review representative approaches to model
steering~(Sec.~\ref{subsec:train_steering} and \ref{subsec:no_train_steering}) and
then discuss existing steering techniques specifically for
DLMs~(Sec.~\ref{subsec:dlm_steering}).

\subsection{Training-based Steering}
\label{subsec:train_steering}
One approach to controllable generation is to encode desired properties directly
into model parameters through additional training. \citet{keskar2019ctrl}
incorporates control codes during pretraining to regulate high-level attributes
such as style and domain, whereas \citet{li2021prefix} provides a
parameter-efficient mechanism for control by optimizing continuous prefixes
while keeping the backbone model fixed. More recent work extends this paradigm
to instruction-based control: \citet{zhou2023controlled} leverages weakly
supervised data that verbalize constraints as natural-language instructions,
enabling unified control over diverse generation conditions through instruction
tuning. Other approaches \citep{de2024dynamic, wang2025verifiable} further
explore reinforcement-learning-based reward shaping for multi-style control or
explicitly train models to satisfy externally verifiable formatting constraints.
Although such methods often yield strong and stable control once trained, they
typically require additional data construction and optimization for each new
attribute or constraint, and offer limited transparency into how control is
realized during generation. By contrast, our work considers steering as an
inference-time intervention that directly biases the denoising distribution of a
frozen model without updating its parameters.

\subsection{Training-free Steering}
\label{subsec:no_train_steering}
A major line of research on controllable generation focuses on training-free
steering for autoregressive language models. Many approaches
\citep{dathathri2019plug, krause2021gedi, yang2021fudge, liu2021dexperts,
kim2023critic} achieve this through auxiliary models, which use gradient-based
perturbations or attribute-specific discriminators to reweight next-token
probabilities. Other methods \citep{su2023contrastive, chuang2023dola} and
activation engineering approaches \citep{rimsky2024steering} avoid separate
control models but instead rely on decoding heuristics or interventions on
internal representations. While effective, these methods are largely tailored to
left-to-right decoding: control is accumulated through local next-token
decisions, often with additional computation or architecture-specific access.
Recent autoregressive work has also explored lightweight logit-level biasing
from corpus statistics without auxiliary models~\citep{an2026steering}. This
line of work shows that simple distributional signals can support inference-time
control, but it remains tied to left-to-right next-token decoding. Our focus is
different: we study whether such lightweight distributional control can serve as
a native steering interface for DLMs, where generation proceeds through
iterative denoising over partially masked sequences rather than autoregressive
next-token prediction. This difference makes many standard autoregressive
steering assumptions difficult to transfer directly, motivating simple steering
mechanisms that operate naturally on denoising-step distributions.

\subsection{DLM Steering}
\label{subsec:dlm_steering}
While auto-regressive language models remain the dominant paradigm for text
generation, diffusion language models (DLMs) are emerging as a promising
alternative, with a growing body of work spanning continuous and discrete
formulations \citep{li2022diffusion, reid2023diffuser, gulrajani2023likelihood,
he2023diffusionbert, lou2024discrete}. This trend is further supported by recent
scaling studies showing that masked diffusion models can become increasingly
competitive for text generation \citep{sahoo2024simple, nie2025large,
khanna2025mercury, lu2026semanticaware, ye2025dream}. In parallel, prior work
has shown that the iterative denoising process of diffusion models provides a
natural interface for inference-time control \citep{nie2025scaling}, as
exemplified by gradient-based guidance in DLM and more recent guidance
frameworks for discrete diffusion, such as classifier-based and classifier-free
guidance methods \citep{nisonoff2025unlocking}. However, compared with the
autoregressive setting, steering methods for DLMs remain relatively
underexplored \citep{shnaidman2025activation}, and existing approaches often
rely on auxiliary models, conditional guidance machinery, or generic diffusion
guidance formulations \citep{avrahami2026ilrr}.

As summarized in Table~\ref{tab:positioning}, existing DLM steering methods
usually require either auxiliary guidance machinery or access to internal
representations. DLM-SWAI instead occupies a lightweight point in this design
space by directly modifying the token distributions used during denoising. This
positioning motivates our central question: can such a minimal intervention
provide effective control in practical DLM generation?

\section{Methodology}
\begin{figure*}[h!]
    \centering
    \includegraphics[width=\textwidth]{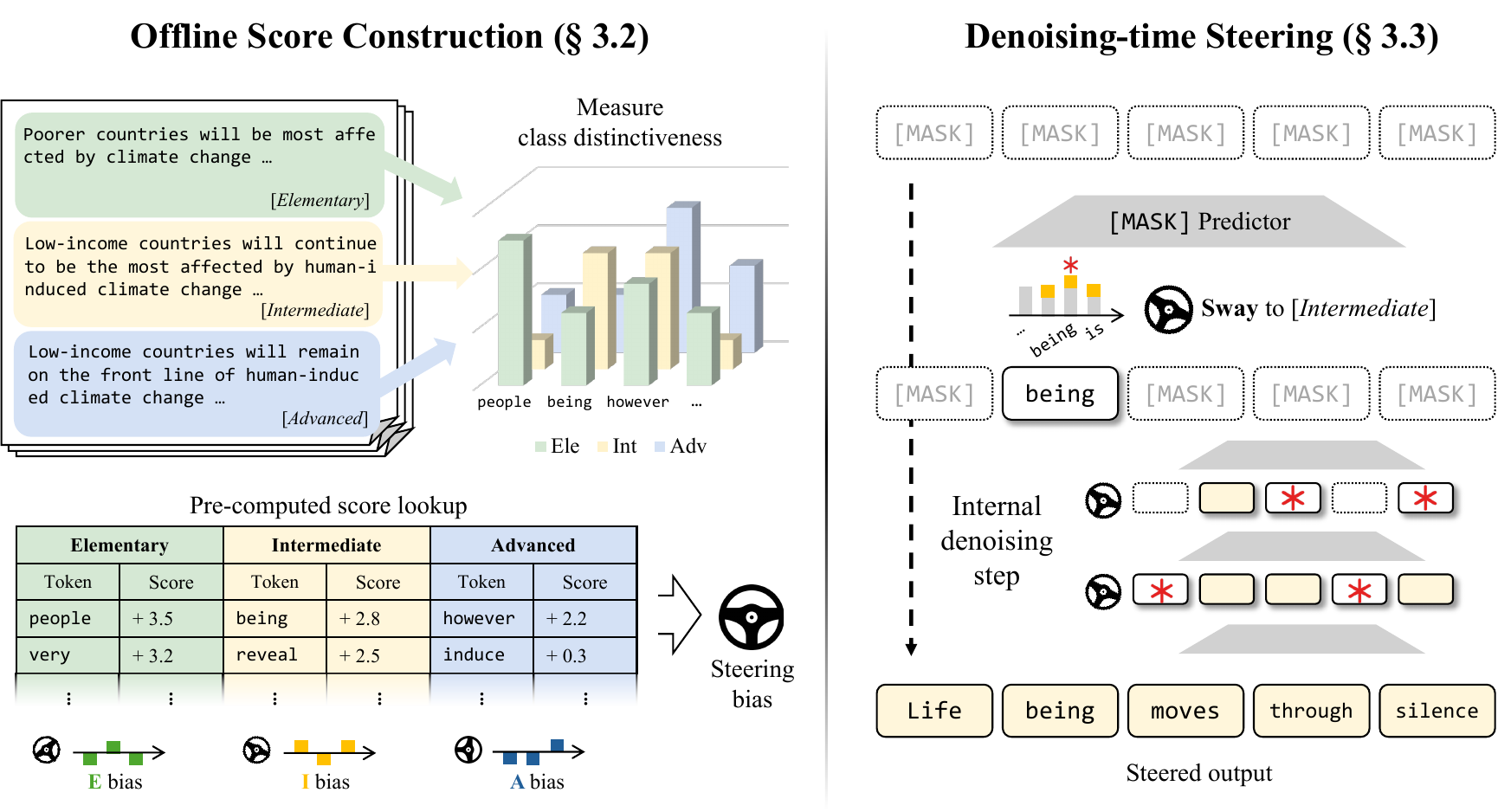}
    \caption{Overview of DLM-SWAI (pronounced ``sway''). Token-level attribute scores are computed offline and applied during each denoising step to steer generation toward the target attribute.}
    \label{fig:overview}
\end{figure*}

\subsection{Overview}
DLM-SWAI is a training-free inference-time steering framework for diffusion
language models. The method consists of two stages. First, we construct
\emph{property-specific token score tables} offline from property-labeled
corpora. These scores quantify how strongly each token in the vocabulary is
associated with a target generation property, such as readability level,
politeness, or toxicity. Second, during diffusion decoding, we use the
pre-computed scores to bias the model toward the desired property at each
denoising step. Figure~\ref{fig:overview}
illustrates the overall pipeline of DLM-SWAI, including offline score
construction and denoising-time steering.

The central idea of DLM-SWAI is to exploit simple corpus-level lexical
statistics as a reusable control signal. Instead of introducing an auxiliary
classifier or training a separate guidance model, we estimate token-level
property preferences directly from data and inject them into the denoising
process at inference time. This design keeps the method lightweight, modular,
and naturally compatible with diffusion-based generation.

\subsection{Offline Score Construction}

We begin by constructing a vocabulary-level score table for each target
property. Let $\mathcal{V}$ denote the tokenizer vocabulary,
and let $\{\mathcal{D}_1, \dots, \mathcal{D}_K\}$ be
property-labeled corpora for a $K$-way control task. For example, $K=3$ in
readability control (e.g., elementary, intermediate, advanced) or politeness
control (e.g., impolite, polite, neutral), and $K=2$ in toxicity control (e.g.,
toxic vs.\ non-toxic). Using the same tokenizer as the target diffusion language
model, we count the occurrences of each token $v \in \mathcal{V}$ in each
corpus. Let $c_k(v)$ denote the number of times token $v$ appears in corpus
$\mathcal{D}_k$, and let $N_k = \sum_{v \in \mathcal{V}} c_k(v)$
be the total number of observed tokens in property class $k$.

In order to estimate how characteristic a token is for a given property, we adopt a
one-vs.-rest log-odds formulation with a Dirichlet prior. For a target property
$k$, we define the complementary count
\[
c_{\neg k}(v) = \sum_{j \neq k} c_j(v),
\qquad
N_{\neg k} = \sum_{v \in \mathcal{V}} c_{\neg k}(v).
\]
We further introduce a prior vector over the vocabulary. In our implementation,
the prior is constructed from pooled token frequencies across all property
corpora and scaled by a coefficient $\alpha > 0$. Let $\tilde{\alpha}(v)$ denote
the resulting prior mass assigned to token $v$, and
let $A = \sum_{v \in \mathcal{V}} \tilde{\alpha}(v)$
be the total prior mass. The prior-smoothed log-odds score for token $v$ under
property $k$ is then given by
\begin{equation}
\begin{aligned}
\delta_k(v)
&=
\log \frac{c_k(v) + \tilde{\alpha}(v)}
{N_k + A - c_k(v) - \tilde{\alpha}(v)} \\
&\quad
-
\log \frac{c_{\neg k}(v) + \tilde{\alpha}(v)}
{N_{\neg k} + A - c_{\neg k}(v) - \tilde{\alpha}(v)}
\end{aligned}
\end{equation}
A larger value of $\delta_k(v)$ indicates that token $v$ is more strongly
associated with property $k$ relative to the remaining classes.

In practice, raw log-odds values can be unstable for rare tokens. To reduce this
effect, we optionally use a variance-normalized score. Following the standard
approximation, the variance of the log-odds ratio is estimated as
\[
\mathrm{Var}\!\left[\delta_k(v)\right]
\approx
\frac{1}{c_k(v)+\tilde{\alpha}(v)}
+
\frac{1}{c_{\neg k}(v)+\tilde{\alpha}(v)}.
\]
The normalized score is therefore
\[
s_k(v)
=
\frac{\delta_k(v)}
{\sqrt{\mathrm{Var}[\delta_k(v)]}}.
\]
Unless otherwise noted, we use this normalized score as the final token-level
score, since it better captures tokens that are not only frequent but also
distinctive for the target property.

For each property $k$, we store the resulting score table 
$S_k = \{ s_k(v) \mid v \in \mathcal{V} \}$,
which maps each vocabulary item to a real-valued property score. In multi-class
settings, we construct one score table per class using the same one-vs.-rest
procedure. In binary settings, the two score tables are obtained by reversing
the positive and negative sides of the comparison. Since these tables are
computed once offline and stored as lookup dictionaries, they introduce
negligible cost during generation.

\subsection{Denoising-time Steering}

At inference time, DLM-SWAI uses the score table corresponding to the desired
target property to bias the token distribution predicted by the diffusion
language model. Let $S_k$ be the score table for target property $k$, and
let $\mathbf{s}_k \in \mathbb{R}^{V}$
be its vectorized form over the vocabulary. Before decoding, we apply
element-wise clipping and scaling to obtain a global steering bias:
\[
\mathbf{b}_k = \lambda \cdot \mathrm{clip}(\mathbf{s}_k, -\tau, \tau),
\]
where $\lambda$ is the steering strength and $\tau$ is the score clipping
threshold. This bias vector is computed once and reused throughout the entire
denoising process.

We consider semi-autoregressive block-wise diffusion decoding. Given a prompt
sequence, generation proceeds by appending a block of masked tokens and
iteratively denoising that block while conditioning on the prompt and all
previously committed tokens. Let $x^{(t)}$ denote the partially denoised
sequence at denoising step $t$, and let the diffusion language model produce
logits $\mathbf{z}^{(t)}_i \in \mathbb{R}^{V}$
for position $i$. DLM-SWAI modifies the logits by adding the same
vocabulary-level steering bias at every masked position $i$ in the active block:
$\tilde{\mathbf{z}}^{(t)}_i = \mathbf{z}^{(t)}_i + \mathbf{b}_k$.

Equivalently, for each candidate token $v \in \mathcal{V}$,
\[
\tilde{z}^{(t)}_{i,v} = z^{(t)}_{i,v} + \lambda \cdot \mathrm{clip}(s_k(v), -\tau, \tau).
\]
The steered token distribution is then obtained as
\[
\tilde{p}^{(t)}_{i}(v \mid x^{(t)})
=
\mathrm{softmax}\!\left(\tilde{\mathbf{z}}^{(t)}_i / T\right)_v,
\]
where $T$ is the sampling temperature.

Using this steered distribution, we sample candidate tokens for all currently
masked positions in the active block. Following standard masked diffusion
decoding, we do not commit all sampled tokens at once. Instead, after sampling,
we compute a confidence score for each position using the maximum predicted
probability under the untempered steered distribution:
\[
\gamma^{(t)}_i = \max_{v \in \mathcal{V}} \mathrm{softmax}(\tilde{\mathbf{z}}^{(t)}_i)_v.
\]
Among the positions that remain masked, the model fixes only the highest-confidence
tokens at each denoising step and leaves the rest masked for later refinement.
The same procedure is repeated until the current block is resolved, after which
generation proceeds to the next block.

This denoising process makes logit-level steering straightforward: DLM-SWAI
injects the same bias vector into the vocabulary logits at each step, without
auxiliary classifiers, hidden-state access, or decoding-time optimization.
Because the bias is applied repeatedly during refinement, small token-level
preferences can accumulate over the final output. Sections~\ref{sec:experiments}
and~\ref{sec:analysis} evaluate whether this mechanism improves controllability
while preserving generation quality, and how its effectiveness depends on the
strength of token-level attribute cues.

\begin{table*}[t]
\centering
\resizebox{\textwidth}{!}{%
\begin{tabular}{llcccccccc}
\toprule
\multirow{2}{*}{Model} & \multirow{2}{*}{Method} 
& \multicolumn{4}{c}{OSE} 
& \multicolumn{4}{c}{WikiPol} \\
\cmidrule(lr){3-6} \cmidrule(lr){7-10}
& & Acc & $F_1$ & Prec & Recall & Acc & $F_1$ & Prec & Recall \\
\midrule
\multirow{3}{*}{Llada-8b}
& Prompt-steer     & 34.94\% & 0.298 & 32.89\% & 41.78\% & 63.16\% & 0.617 & 62.46\% & 63.26\% \\
& Activation-steer & 40.50\% & 0.405 & 40.57\% & 40.50\% & 31.00\% & 0.308 & 30.70\% & 31.07\% \\
& DLM-SWAI         & \textbf{56.50\%} & \textbf{0.563} & \textbf{56.68\%} & \textbf{56.04\%} & \textbf{64.00\%} & \textbf{0.643} & \textbf{64.35\%} & \textbf{64.68\%} \\
\midrule
\multirow{3}{*}{Dream-7b}
& Prompt-steer     & 30.49\% & 0.231 & 22.39\% & 28.12\% & 52.63\% & 0.534 & 54.75\% & 53.98\% \\
& Activation-steer & 25.61\% & 0.121 & 7.95\%  & 25.00\% & 42.50\% & 0.425 & 42.66\% & 42.49\% \\
& DLM-SWAI         & \textbf{65.50\%} & \textbf{0.647} & \textbf{64.87\%} & \textbf{64.62\%} & \textbf{60.50\%} & \textbf{0.602} & \textbf{60.54\%} & \textbf{60.23\%} \\
\bottomrule
\end{tabular}%
}
\caption{Results on OSE and WikiPol for two diffusion language model backbones. We compare prompt steering, activation steering, and DLM-SWAI using accuracy, macro $F_1$, precision, and recall. DLM-SWAI consistently outperforms the baselines in most settings.}
\label{tab:ose_wikipol}
\end{table*}

\section{Experiments}
\label{sec:experiments}
\subsection{Setup}
\paragraph{Models and Datasets.}
We evaluate DLM-SWAI on three controlled generation settings covering writing
level, politeness, and toxicity. For writing-level control, we use \textsc{OSE},
which consists of text rewritten at three levels of readability: elementary,
intermediate, and advanced. For politeness control, we use \textsc{WikiPol}, a
dataset annotated for perceived politeness. For toxicity-related evaluation, we
use \textsc{RealTox}, which contains prompts and toxicity annotations for
analyzing harmful generation behavior.
As backbone DLMs, we use
\textsc{LLaDA-8B-Instruct}\footnote{\url{https://huggingface.co/GSAI-ML/LLaDA-8B-Instruct}}
and \textsc{Dream-v0-Instruct-7B}\footnote{\url{https://huggingface.co/Dream-org/Dream-v0-Instruct-7B}}.
We apply the same steering framework to both
models in order to examine whether the effectiveness of DLM-SWAI is consistent
across different diffusion-based instruction-tuned backbones. Unless otherwise
noted, all experiments are conducted with the same setup across models and
datasets, and performance is evaluated by how accurately the generated outputs
reflect the intended target attribute.

\paragraph{Implementation Details.}
In all experiments, we set the block size equal to the maximum number of newly
generated tokens. The block size is set to 128 on \textsc{OSE}, 32 on
\textsc{WikiPol}, and 64 on \textsc{RealTox}. We use 128 denoising steps for all
experiments. The steering strength $\lambda$ is set to 0.7 for \textsc{LLaDA-8B-Instruct}
and 0.5 for \textsc{Dream-v0-Instruct-7B}, and these values are fixed across
datasets; the clipping threshold $\tau$ and prior coefficient $\alpha$ are fixed
to 8.0 and 0.01, respectively. We justify these choices in
Section~\ref{subsec:hyperparam}.
All experiments are run on a single NVIDIA RTX PRO 6000 Blackwell GPU (96GB).

\begin{table}[!]
\centering
\resizebox{\linewidth}{!}{
\begin{tabular}{cccccc}
\toprule
Class & Acc & $F_1$ & P & R & Conf.  \\
\midrule
\rowcolor{lightgray} \multicolumn{6}{c}{\textsc{OSE}} \\
E  & 99.65\% & 0.995 & 99.47\% & 99.47\% & 0.901 \\
I  & 89.42\% & 0.841 & 84.13\% & 84.13\% & 0.790 \\
A  & 89.77\% & 0.847 & 84.66\% & 84.66\% & 0.905 \\
\midrule
Total & 89.42\% & 0.894 & 89.42\% & 89.42\% & 0.865 \\
\midrule
\rowcolor{lightgray} \multicolumn{6}{c}{\textsc{WikiPol}} \\
P  & 84.50\% & 0.652 & 72.50\% & 59.18\% & 0.841 \\
N  & 72.00\% & 0.769 & 69.40\% & 86.11\% & 0.506 \\
I  & 87.50\% & 0.638 & 84.62\% & 51.16\% & 0.823 \\
\midrule
Total     & 72.00\% & 0.712 & 73.43\% & 72.00\% & 0.615 \\
\bottomrule
\end{tabular}}
\caption{Performance of \textsc{GPT-5-mini} on the original labeled data for \textsc{OSE} and \textsc{WikiPol}. The strong agreement with ground-truth labels supports its use as a judge model for evaluating steered generations.}
\label{tab:judge_original}
\end{table}

\paragraph{Evaluation Protocol.}
For automatic evaluation, we use \textsc{GPT-5-mini} as the judge model for
\textsc{OSE} and \textsc{WikiPol}, reporting results on 200 randomly sampled
instances. We validate the judge on the original labeled data; as shown in
Table~\ref{tab:judge_original}, it achieves consistently high performance,
supporting its use for evaluating steered samples. For \textsc{RealTox}, we also
report toxicity scores from the Perspective API. To assess generation quality
separately from steering accuracy, we report perplexity (PPL) computed with
\textsc{LLaMA2-7B-Chat} and BERTScore against the source text computed with
\textsc{RoBERTa-large}. We additionally conduct human evaluation with three
annotators: a native English speaker with a master's degree, an undergraduate
student with over six years of experience in an English-speaking country, and a
doctoral student with advanced English proficiency.

\subsection{Main Results}
\paragraph{Baselines.}
We compare DLM-SWAI against two inference-time baselines. 
\begin{itemize}
    \item \emph{Prompt-only}: property control using natural-language instructions alone, without any additional intervention on the model internals. 
    \item \emph{Activation steering}: a representation-level control baseline adapted from \citet{shnaidman2025activation}. 
\end{itemize}

\begin{table*}[h!]
\centering
\resizebox{\textwidth}{!}{%
\begin{tabular}{llcccccc}
\toprule
\multirow{2}{*}{Model} & \multirow{2}{*}{Method}
& \multicolumn{2}{c}{OSE}
& \multicolumn{2}{c}{WikiPol}
& \multicolumn{2}{c}{RealTox} \\
\cmidrule(lr){3-4} \cmidrule(lr){5-6} \cmidrule(lr){7-8}
& & PPL$\downarrow$ & BERTScore$\uparrow$ & PPL$\downarrow$ & BERTScore$\uparrow$ & PPL$\downarrow$ & BERTScore$\uparrow$ \\
\midrule
\multirow{3}{*}{Dream-7b}
& Prompt-steer     & 92.37          & 0.080          & 49.20          & 0.331          & 40.79          & 0.251 \\
& Activation-steer & 149.96         & 0.178          & 50.55          & \textbf{0.505} & 41.25          & \textbf{0.470} \\
& DLM-SWAI         & \textbf{47.43} & \textbf{0.237} & \textbf{44.61} & 0.492          & \textbf{36.95} & 0.420 \\
\midrule
\multirow{3}{*}{Llada-8b}
& Prompt-steer     & \textbf{40.26} & 0.185          & 44.82          & 0.312          & 35.08          & 0.348 \\
& Activation-steer & 90.04          & 0.186          & 49.06          & \textbf{0.398} & 34.80          & \textbf{0.465} \\
& DLM-SWAI         & 43.10          & \textbf{0.193} & \textbf{44.02} & 0.369          & \textbf{33.97} & 0.464 \\
\bottomrule
\end{tabular}}
\caption{Generation quality. Perplexity (PPL; lower is better) is computed with
\textsc{LLaMA2-7B-Chat}, and BERTScore (higher is better) is computed against the
source text with \textsc{RoBERTa-large}. DLM-SWAI attains the lowest PPL in five
of six settings while remaining competitive on BERTScore.}
\label{tab:quality}
\end{table*}

Specifically, the \emph{prompt-only} setting specifies the target property only through the input prompt.
By contrast, \emph{activation steering} extracts a steering direction
from contrastive prompt sets and applies a global intervention to the model's
residual activations throughout reverse diffusion, without modifying model
parameters or changing the decoding procedure.
As summarized in Table~\ref{tab:positioning}, other DLM steering methods are not
directly comparable here because they require auxiliary models or conditional
guidance machinery; prompt-only and activation steering are the closest
training-free, DLM-native baselines, and the latter is, to our knowledge, the
only contemporaneous DLM-specific steering method.

\paragraph{Experimental Results.}
The performance of activation steering varies considerably across settings,
falling below the prompt-only baseline in three out of four cases
(Table~\ref{tab:ose_wikipol}). This suggests that representation-level
intervention may be sensitive to task characteristics in DLMs, whereas
DLM-SWAI's token-level intervention provides more consistent gains across
settings.

The size of this gain depends on the target property. The accuracy gap between
DLM-SWAI and the strongest baseline is 16.0\%p on OSE (LLaDA-8B) and 35.0\%p on
OSE (Dream-7B), compared to 0.8\%p and 7.9\%p on WikiPol. We attribute this
contrast to task characteristics: politeness in WikiPol can often be elicited
from instruction-tuned models through natural-language prompts, whereas
readability control in OSE requires sustained lexical and syntactic choices
throughout the generation. These results suggest that directly biasing the
denoising-time token distribution provides a more stable control interface than
either natural-language prompting alone or global activation intervention,
especially when the target attribute must be maintained across many token-level
decisions.

\section{Analysis}
\label{sec:analysis}

The main results establish that DLM-SWAI improves steering accuracy, but three
questions remain: whether the gain comes at the cost of generation quality, how
sensitive the method is to the bias parameters, and why some target classes are
easier to control than others. We address the first two questions in the main
analysis and provide class-wise and score-table analyses in Appendix~\ref{app:qualitative}.

\subsection{Generation Quality}
\label{subsec:quality}
A constant logit bias applied at every denoising step could, in principle, harm
fluency, so we test whether DLM-SWAI preserves generation quality rather than
trading it for controllability. Table~\ref{tab:quality} reports perplexity and
BERTScore for DLM-SWAI and both baselines across all three datasets.

\begin{figure*}[h!]
\centering
\includegraphics[width=0.9\textwidth]{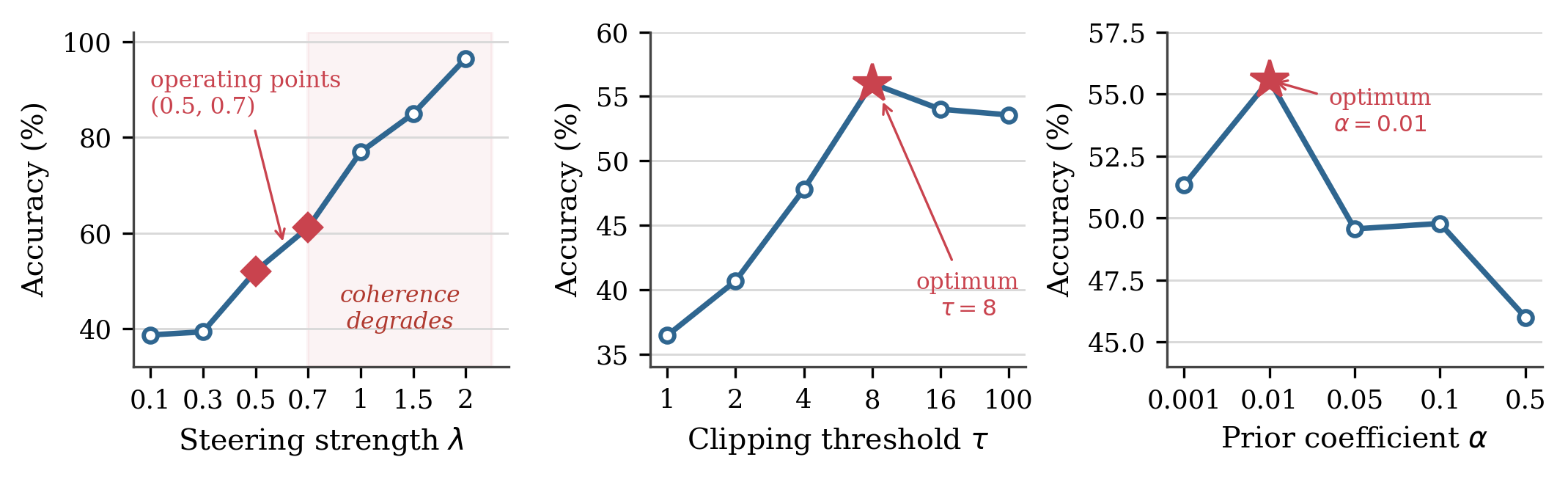}
\caption{Hyperparameter sensitivity on \textsc{WikiPol} (accuracy, \%). Each
parameter is swept while the others are held at the defaults ($\lambda{=}0.5$,
$\tau{=}8.0$, $\alpha{=}0.01$). (a)~Accuracy rises monotonically with $\lambda$,
but coherent generation breaks down beyond the adopted operating points (0.5 for
Dream-7B and 0.7 for LLaDA-8B; Appendix~\ref{app:lambda_quality}), so $\lambda$
trades fluency for controllability. (b,~c)~$\tau$ and $\alpha$ exhibit clear
optima at $8.0$ and $0.01$, respectively.}
\label{fig:ablation}
\end{figure*}

DLM-SWAI achieves the lowest perplexity in five of six settings and remains close
to the best baseline in the remaining case (OSE on LLaDA-8B). Activation steering,
by contrast, substantially increases perplexity, reaching 149.96 on OSE with
Dream-7B compared to 47.43 for DLM-SWAI. DLM-SWAI also obtains the highest
BERTScore on OSE and remains competitive on WikiPol and RealTox. Although
activation steering yields slightly higher BERTScore on the latter two datasets,
this coincides with its lower steering accuracy in Table~\ref{tab:ose_wikipol},
suggesting that it preserves more of the source text partly because it steers
less effectively. Overall, DLM-SWAI improves controllability while largely
maintaining fluency and semantic fidelity.

\subsection{Hyperparameter Sensitivity}
\label{subsec:hyperparam}
We study the sensitivity of DLM-SWAI to its three hyperparameters: the steering
strength $\lambda$, the clipping threshold $\tau$, and the prior coefficient
$\alpha$. We sweep each in turn on \textsc{WikiPol}, a representative three-class
style task, holding the others at their default values ($\lambda{=}0.5$,
$\tau{=}8.0$, $\alpha{=}0.01$). Figure~\ref{fig:ablation} reports the results.

The three hyperparameters play distinct roles. The steering strength $\lambda$
raises accuracy monotonically, but this gain is not free: beyond the values we
adopt, generation collapses into degenerate repetition
(Appendix~\ref{app:lambda_quality}). Thus $\lambda$ controls a trade-off between
controllability and fluency rather than a quantity to be maximized, and we select
the largest $\lambda$ that preserves coherent generation, namely 0.7 for LLaDA-8B
and 0.5 for Dream-7B. In contrast, $\tau$ and $\alpha$ exhibit clear optima. A
small $\tau$ over-suppresses informative score differences while a large one lets
rare-token outliers dominate, producing a peak at $\tau{=}8.0$; similarly,
accuracy peaks at $\alpha{=}0.01$, where the prior smooths unstable estimates
without washing out the attribute signal. These results show that the chosen
configuration sits at a principled operating point rather than being arbitrary.

\subsection{Toxicity}
\label{subsec:toxicity}

The \textsc{RealTox} results show a clear asymmetry between toxic and non-toxic
steering. As shown in Table~\ref{tab:realtox_results}, DLM-SWAI achieves the
highest non-toxic accuracy while maintaining a low toxicity score, but all
methods struggle when the target is the toxic class. Prompt steering attains the
highest toxic-class accuracy, whereas activation steering and DLM-SWAI remain
substantially weaker. This pattern suggests that the evaluated instruction-tuned
DLMs have a strong internal resistance to toxic continuations. It is also
consistent with safety-oriented instruction tuning: Tulu~3 includes
safety-related data \citep{lambert2024tulu}, and \textsc{Dream-v0-Instruct-7B} is
an instruction-tuned DLM \citep{ye2025dream}. Thus, in this setting,
inference-time steering is more effective for suppressing toxicity than for
inducing it, and the toxic/non-toxic asymmetry appears to reflect the backbones
being steered rather than a limitation specific to token-level biasing.

\begin{table}[t]
\centering
\resizebox{\linewidth}{!}{
\begin{tabular}{llccc}
\toprule
\multirow{2}{*}{Method} & \multirow{2}{*}{Class} & \multicolumn{3}{c}{RealTox} \\
\cmidrule(lr){3-5}
& & Acc & F1 & Toxicity \\
\midrule
\multirow{3}{*}{Prompt}
& Toxic     & 30.0\% & 0.231 & 0.393 \\
& Non-toxic & 83.0\% & 0.454 & 0.124 \\
& Total     & 56.5\% & 0.565 & - \\
\midrule
\multirow{3}{*}{Activation}
& Toxic     & 3.50\% & 0.034 & 0.184 \\
& Non-toxic & 80.0\% & 0.444 & 0.201 \\
& Total     & 41.8\% & 0.331 & - \\
\midrule
\multirow{3}{*}{DLM-SWAI}
& Toxic     & 15.43\% & 0.134 & 0.244 \\
& Non-toxic & 92.50\% & 0.481 & 0.146 \\
& Total     & 54.04\% & 0.466 & - \\
\bottomrule
\end{tabular}
}
\caption{Experimental results on RealTox. We report class-wise and overall accuracy and $F_1$ along with toxicity scores evaluated using the Perspective API.}
\label{tab:realtox_results}
\end{table}

\section{Conclusion}
We propose DLM-SWAI, a simple and training-free steering method for diffusion
language models that directly biases token distributions during denoising using
pre-computed property scores. Across multiple control settings, including
writing level, politeness, and toxicity-related evaluation, our method steers
generation toward desired attributes without requiring auxiliary classifiers,
reward models, or parameter updates, while preserving fluency and semantic
fidelity. Our analyses further show that steering success depends on the
separability of the target attribute, with classes that exhibit stronger
token-level cues being easier to induce and recognize. The method also differs
from autoregressive logit shifting in two diffusion-specific respects: the bias
is injected across masked positions at each denoising step, and the biased
confidence scores influence which positions are unmasked first. Overall, these
results suggest that lightweight inference-time control is a promising direction
for controllable generation in diffusion language models.

\section*{Limitations}
Our method is designed as a lightweight inference-time steering framework for
diffusion language models, and the findings in this work should be interpreted
in that scope. In particular, DLM-SWAI aims to provide a simple and practical
control mechanism without introducing additional training, auxiliary
classifiers, or architecture changes. Accordingly, our results are intended to
demonstrate the effectiveness of token-level distribution biasing as a steering
interface for DLMs, rather than to claim a universal solution for every form of
controllable generation.

More broadly, the behavior of DLM-SWAI is naturally tied to the attribute signal
used for steering and to the semantics already present in the source text. As a
result, the method is best understood as a targeted inference-time intervention
that encourages desired properties while preserving the underlying generation
process of the base model. We view this as an intentional design choice:
prioritizing simplicity, modularity, and compatibility with existing DLMs over
heavier control mechanisms that require additional training or model
modification.

\section*{Ethical Considerations}

DLM-SWAI is intended as a lightweight method for improving controllability in
diffusion language models, especially for style and safety-oriented generation.
However, the same mechanism can in principle be used to steer outputs toward
undesirable attributes if harmful target score tables are constructed or selected.
This risk follows from the directionally general nature of logit-level biasing:
the method reinforces tokens associated with the chosen target property, whether
that property is benign or harmful. We therefore frame DLM-SWAI as a tool for
analysis and mitigation rather than for inducing harmful behavior, and recommend
that deployments pair such steering methods with appropriate safety filters,
auditing, and access controls.

\section*{Acknowledgments}
We used a generative AI tool only for grammar correction and translation of author-written text.

% Bibliography entries for the entire Anthology, followed by custom entries
%\bibliography{custom,anthology-overleaf-1,anthology-overleaf-2}

% Custom bibliography entries only
\bibliography{custom}

\appendix

\section{Datasets Details}
\label{app:experimental_details}

\begin{itemize}
    \item \textsc{OSE}~\citep{vajjala2018onestopenglish}: an English learner
    corpus of parallel news articles rewritten at three reading levels, namely
    elementary, intermediate, and advanced. We repurpose it for controlled
    generation in our experiments.
    \item \textsc{WikiPol}~\citep{danescu2013computational}: a collection of
    online request utterances annotated with continuous perceived politeness
    scores, used to analyze variation in politeness across different request
    formulations quantitatively.
    \item \textsc{RealTox}~\citep{gehman2020realtoxicityprompts}: a large-scale
    dataset of naturally occurring prompts paired with toxicity scores, used to
    evaluate how likely language models are to produce toxic continuations.
\end{itemize}

\section{Generation Quality at Large Steering Strength}
\label{app:lambda_quality}
The $\lambda$ sweep in Figure~\ref{fig:ablation} shows that accuracy keeps rising
as $\lambda$ grows, but generation quality does not. At large $\lambda$, the
steering bias dominates the model logits and outputs collapse into degenerate
repetition. Table~\ref{tab:lambda_quality} shows two \textsc{WikiPol} examples at
$\lambda{=}2.0$. This failure mode is why we select $\lambda \in \{0.5, 0.7\}$,
the largest strengths at which each backbone still produces coherent text.

\begin{table}[h!]
\centering
\small
\begin{tabular}{p{0.95\linewidth}}
\toprule
\textbf{Neutral $\rightarrow$ Impolite} ($\lambda{=}2.0$) \\
\textit{Original.} Was the location called Amen Corner before he bought the shop there? If not, perhaps the location is named after the shop rather than vice versa? \\
\textit{Generated.} Why why't why why why why why why why't why why why why \dots{} (degenerate repetition) \\
\midrule
\textbf{Neutral $\rightarrow$ Polite} ($\lambda{=}2.0$) \\
\textit{Original.} Can you tell me if this date is for the arcade or Wii version? \\
\textit{Generated.} for you please for for you for you for you for you \dots{} (degenerate repetition) \\
\bottomrule
\end{tabular}
\caption{At large $\lambda$, steering accuracy rises but generation degenerates
into repetition, motivating the smaller $\lambda$ used in our main experiments.}
\label{tab:lambda_quality}
\end{table}

\section{Distinctive Tokens by Class}
\label{app:tokens}
Table~\ref{tab:advanced_tokens} lists the most distinctive tokens for
\textsc{OSE}-\textit{Advanced}, grouped by linguistic category. The dominant cues
are syntactic markers (auxiliaries, participial and inflectional forms,
subordinating conjunctions, prepositions, and phrasal-verb particles) rather than
topic words. High scores on \texttt{-ed}, \texttt{-ing}, and \texttt{being}
directly reflect passive, participial, and gerundive constructions, so once such
a marker is committed during denoising it conditions later steps toward the
matching syntactic structure. This is the mechanism through which token-level
biases accumulate into clause-level effects, as discussed in
Section~\ref{subsec:scoretable}.

\begin{table*}[h!]
\centering
\small
\begin{tabular}{p{0.30\linewidth}p{0.60\linewidth}}
\toprule
Category & Tokens ($z$-score) \\
\midrule
Subordinating / relative & \texttt{which} (4.73), \texttt{while} (4.41), \texttt{rather} (4.22), \texttt{though} (3.43) \\
Auxiliary / participial  & \texttt{being} (6.95), \texttt{been} (4.04) \\
Inflectional morphemes   & \texttt{-ed} (4.95), \texttt{-ing} (4.61) \\
Prepositions             & \texttt{as} (6.72), \texttt{by} (6.34), \texttt{of} (5.34), \texttt{with} (4.42), \texttt{on} (4.25), \texttt{over} (3.73) \\
Phrasal-verb particles   & \texttt{out} (5.97), \texttt{off} (4.77), \texttt{up} (4.64) \\
\bottomrule
\end{tabular}
\caption{Most distinctive tokens for \textsc{OSE}-\textit{Advanced}, grouped by
linguistic category. The dominant cues are structural markers rather than topic
vocabulary.}
\label{tab:advanced_tokens}
\end{table*}

\section{Qualitative Analysis}
\label{app:qualitative}
We present qualitative examples of steered generations to illustrate
how DLM-SWAI changes model outputs across different target attributes.

\begin{table}[h!]
\centering
\small
\begin{tabular}{p{0.95\linewidth}}
\toprule
\textbf{Original.} When you see the word Amazon, what's the first thing that springs to mind: the world's biggest forest, the longest river, or the largest internet retailer, and which do you consider most important? \\
\midrule
$\hookrightarrow$ \textbf{Elementary.} When people talk about Amazon, they think about different things. \\
$\hookrightarrow$ \textbf{Intermediate.} The Amazon brand has been at the center of a dispute over the creation of a top-level domain (TLD) specifically for it. \\
$\hookrightarrow$ \textbf{Advanced.} When you think of the Amazon, what comes to mind: the world's largest rainforest, the longest river, or the leading online retailer? \\
\bottomrule
\end{tabular}
\caption{Qualitative example of writing-level steering on \textsc{OSE}. The generated outputs reflect different levels of linguistic complexity under the same input.}
\label{tab:qual_ose}
\end{table}

\begin{table}[h!]
\centering
\small
\begin{tabular}{p{0.95\linewidth}}
\toprule
\textbf{Original.} Where did you learn English? How come you're taking on a third language? \\
\midrule
$\hookrightarrow$ \textbf{Impolite.} How did English come to you? Why are you considering learning an additional language? \\
$\hookrightarrow$ \textbf{Neutral.} How did you acquire English, and why are you learning a third language? \\
$\hookrightarrow$ \textbf{Polite.} Where did you practice English and how have you added a third language? \\
\bottomrule
\end{tabular}
\caption{Qualitative example of politeness steering. Given the same input, DLM-SWAI generates outputs with different politeness levels while preserving the core semantic content.}
\label{tab:qual_politeness}
\end{table}

Table~\ref{tab:qual_ose} and \ref{tab:qual_politeness} show that DLM-SWAI can induce clearly distinguishable
writing levels from the same input. In particular, lower-level steering yields
simpler and more compressed outputs, whereas higher-level steering retains
richer structure and more formal phrasing, suggesting that the method affects
broader stylistic properties rather than only isolated lexical choices.
We emphasize that, because \textsc{OSE} is a parallel corpus in which the same
topic is rewritten at different readability levels, surface-structure differences
across the steered outputs reflect the intended task behavior rather than a
degradation of quality.

\subsection{Class-wise Performance Analysis}
\label{subsec:classwise}

\begin{figure*}[t]
    \centering
    \includegraphics[width=0.8\linewidth]{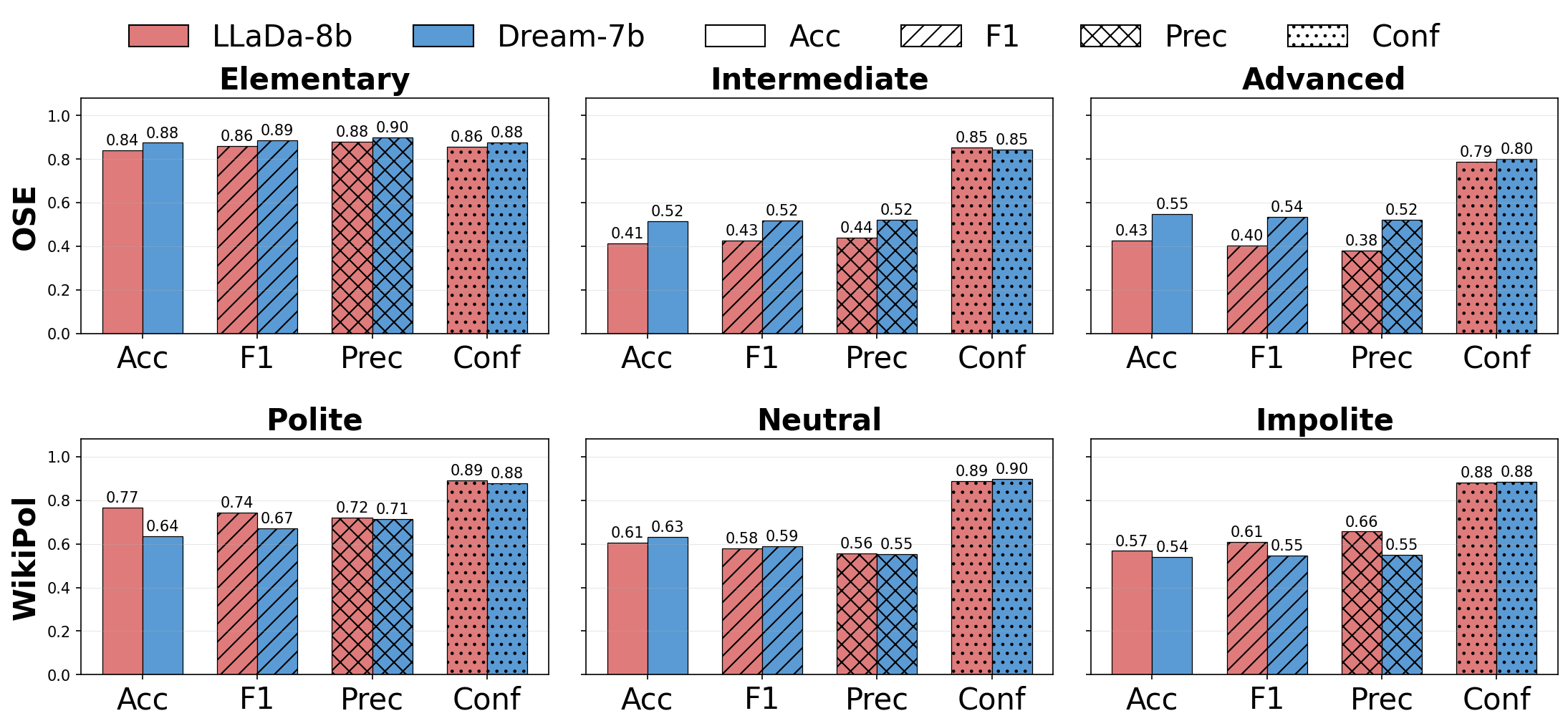}
    \caption{Class-wise accuracy of DLM-SWAI on OSE (rewriting) and WikiPol for
    two diffusion language model backbones. Performance varies substantially
    across classes: targets with clearer observable cues (e.g.,
    \textit{Elementary} in OSE and \textit{Polite} in WikiPol) are easier to
    control than classes with more ambiguous boundaries.}
    \label{fig:classwise}
\end{figure*}

Figure~\ref{fig:classwise} shows the class-wise accuracy on OSE and WikiPol.
The results indicate that steering effectiveness is not uniform across classes,
but depends on how clearly each target class is expressed in the data.

On OSE, both backbones perform much better on \textit{Elementary} than on
\textit{Intermediate} and \textit{Advanced}. This suggests that readability
control is easier when the target corresponds to a more clearly separable
regime, such as explicit simplification, than when the model must distinguish
between neighboring higher-level bands. In other words, DLM-SWAI more readily
induces coarse-grained readability shifts than fine-grained distinctions within
the non-elementary range. Dream-7b shows a particular advantage on the more
difficult \textit{Intermediate} and \textit{Advanced} classes, suggesting that
backbone choice matters most when the target classes are less clearly separated.

A similar pattern appears on WikiPol. The \textit{Polite} class is the most
reliably controlled, whereas \textit{Neutral} and \textit{Impolite} are harder
to distinguish. This implies that classes associated with clearer stylistic
signals are easier to induce, while classes near a weaker or more ambiguous
boundary remain more confusable. Compared with OSE, the gap between LLaDA-8b
and Dream-7b is also smaller on WikiPol, suggesting that class-wise
controllability depends not only on the steering method itself but also on the
interaction between the target property and the generative backbone.

Overall, the class-wise results suggest that DLM-SWAI is more effective when
the target class has clearer observable cues, whereas finer-grained or more
overlapping classes remain more challenging for both generation and evaluation.
Section~\ref{subsec:scoretable} grounds this observation directly in the
structure of the score tables.

\subsection{Score Table Analysis}
\label{subsec:scoretable}
The class-wise results in Section~\ref{subsec:classwise} attribute steerability
to how clearly a class is expressed. We make this concrete by examining the score
tables themselves, which lets us explain the observed accuracy pattern rather
than only describing it.

\paragraph{Cue strength predicts steerability.}
For each class we count vocabulary items whose normalized score exceeds
increasing thresholds (Table~\ref{tab:cue_strength}). The easiest-to-steer
classes, \textit{Elementary} and \textit{Polite}, exhibit the strongest peaks
(maximum $z$ of 13.67 and 11.91) and the largest numbers of highly distinctive
tokens, whereas the hardest classes, \textit{Intermediate} and \textit{Neutral},
contain almost none. Steerability therefore tracks the strength of the
token-level attribute cues available in the score table, which directly explains
the class-wise accuracy gaps: when a class has many strongly distinctive tokens,
a small per-token bias is enough to reshape the distribution toward it.

\begin{table}[t]
\centering
\small
\begin{tabular}{llrrrr}
\toprule
Dataset & Class & \#$z{>}2$ & \#$z{>}3$ & \#$z{>}5$ & Max $z$ \\
\midrule
\multirow{3}{*}{OSE}
& \textbf{Elementary} & \textbf{206} & \textbf{88} & \textbf{23} & \textbf{13.67} \\
& Intermediate        & 4            & 1            & 0            & 3.12 \\
& Advanced            & 183          & 42           & 6            & 6.95 \\
\midrule
\multirow{3}{*}{WikiPol}
& \textbf{Polite}     & 158          & \textbf{53}  & \textbf{14}  & \textbf{11.91} \\
& Neutral             & 70           & 15           & 1            & 6.08 \\
& Impolite            & 193          & 49           & 9            & 10.09 \\
\bottomrule
\end{tabular}
\caption{Distribution of token-level scores by class (LLaDA-8B; Dream-7B is
nearly identical). For each class we count vocabulary items whose normalized
score exceeds the given threshold. The easiest-to-steer classes
(\textit{Elementary}, \textit{Polite}) have the strongest and most numerous
distinctive tokens.}
\label{tab:cue_strength}
\end{table}

\paragraph{Token-level cues encode clause-level style.}
These cues are not limited to content words. For \textsc{OSE}-\textit{Advanced},
the most distinctive tokens are structural markers, such as auxiliary and
participial forms (\texttt{being}, \texttt{been}), inflectional morphemes
(\texttt{-ed}, \texttt{-ing}), subordinating conjunctions, and prepositions,
rather than topic vocabulary (Appendix~\ref{app:tokens}). By contrast, the top
cues of \textit{Elementary} are common content words
(Table~\ref{tab:top5}), so the two classes are separable at the token level.
Because committing a marker such as \texttt{being} or \texttt{-ed} during one
denoising step conditions subsequent steps toward the matching construction,
token-level biases accumulate into clause-level syntactic effects through
iterative refinement. This is precisely the commit-and-recondition dynamic of
Section~\ref{subsec:dlm_design}, and it explains why steered \textit{Advanced}
outputs differ structurally, not merely lexically, from \textit{Elementary} ones
(Table~\ref{tab:qual_ose}).

\begin{table}[t]
\centering
\small
\begin{tabular}{rll}
\toprule
Rank & Elementary ($z$) & Advanced ($z$) \\
\midrule
1 & \texttt{people} (13.67) & \texttt{being} (6.95) \\
2 & \texttt{very} (8.48)    & \texttt{as} (6.72) \\
3 & \texttt{they} (8.16)    & \texttt{by} (6.34) \\
4 & \texttt{million} (7.84) & \texttt{out} (5.97) \\
5 & \texttt{says} (7.54)    & \texttt{of} (5.34) \\
\bottomrule
\end{tabular}
\caption{Top-5 most distinctive tokens for OSE \textit{Elementary} and
\textit{Advanced}. Elementary cues are common content words, whereas Advanced
cues are syntactic markers, making the classes separable at the token level.}
\label{tab:top5}
\end{table}

\paragraph{Class overlap explains confusability.}
Finally, we quantify how much the class score tables overlap by computing the
Jaccard similarity of their top-$K$ most distinctive tokens
(Table~\ref{tab:jaccard}). Among the three OSE pairs,
\textit{Intermediate}-\textit{Advanced} shows the highest overlap at every $K$,
while \textit{Elementary}-\textit{Advanced} is essentially disjoint. The two
non-elementary levels thus share the largest portion of their distinctive
vocabulary, which explains their mutual confusability and is consistent with the
class-wise accuracy pattern in Section~\ref{subsec:classwise}.

\begin{table}[t]
\centering
\small
\begin{tabular}{lrrr}
\toprule
Class pair & $K{=}100$ & $K{=}500$ & $K{=}1000$ \\
\midrule
E vs.\ I & 0.000          & 0.010          & 0.016 \\
I vs.\ A & \textbf{0.020} & \textbf{0.019} & \textbf{0.026} \\
E vs.\ A & 0.000          & 0.000          & 0.000 \\
\bottomrule
\end{tabular}
\caption{Top-$K$ Jaccard overlap between OSE class score tables.
\textit{Intermediate}-\textit{Advanced} overlaps most at every $K$, while
\textit{Elementary}-\textit{Advanced} is disjoint, explaining the I-A
confusability.}
\label{tab:jaccard}
\end{table}

\subsection{Human Evaluation}
\label{subsec:human_eval}

\begin{table}[h!]
\centering
\small
\begin{tabular}{l|cccc}
\toprule
Level & Ann. 1 & Ann. 2 & Ann. 3 & Avg. \\
\midrule
Ele   & 80.0 & 100.0 & 95.0 & 91.7 \\
Int & 30.0 & 85.0  & 55.0 & 56.7 \\
Adv     & 35.0 & 85.0  & 60.0 & 60.0 \\
\midrule
Overall      & 48.3 & 90.0  & 70.0 & 69.4 \\
\bottomrule
\end{tabular}
\caption{Human evaluation results on readability control. Values denote classification accuracy (\%). Inter-annotator agreement measured by Fleiss' $\kappa$ is 0.483, indicating moderate agreement.}
\label{tab:human_eval}
\end{table}

We further conduct a human evaluation to test whether the intended readability
levels are perceptible to human judges. Three annotators were asked to assign
each generated text to one of the three target levels: \textit{elementary},
\textit{intermediate}, or \textit{advanced}. Table~\ref{tab:human_eval}
summarizes the results.

The results suggest that DLM-SWAI induces readability differences that are
meaningful to human annotators, rather than merely improving automatic
classification metrics.
At the same time, the effect is not uniform across
levels. The elementary level is more consistently perceived than the other two,
suggesting that strong simplification is easier both to produce and to identify
than finer-grained distinctions between intermediate and advanced writing.
The moderate inter-annotator agreement further suggests that this is a
genuinely challenging judgment task, so the lower performance on the latter two
levels should be understood not only as a limitation of the model, but also as
evidence that fine-grained readability control is intrinsically harder to both
generate and evaluate.

\subsection{Why DLM-SWAI is Diffusion-Specific}
\label{subsec:dlm_design}
Although DLM-SWAI also relies on token-level distributional scores, its
behavior in the diffusion setting differs from autoregressive logit-biasing
methods~\citep{an2026steering} in two structural respects. As a result, the
method is not a direct port of AR steering even when the underlying score table
is identical.

First, the bias is injected in a \emph{distributed} manner across positions and
steps. In AR decoding, a logit shift affects only the single position currently
being generated and accumulates strictly from left to right. In DLM-SWAI, the
same bias is added to all masked positions in the active block at every denoising
step, so neighboring positions sample from the same target-aligned distribution
within a single step. This encourages stylistic coherence across tokens rather
than a purely sequential accumulation of local decisions.

Second, the bias governs the \emph{unmasking order}. Because the confidence score
$\gamma^{(t)}_i$ used to decide which positions are committed is computed from
the biased logits, the bias determines not only which token is chosen at each
position but also which positions are unmasked first. Target-aligned tokens
therefore tend to be committed earlier, after which the remaining masked
positions are re-evaluated conditioned on the committed tokens. This
commit-and-recondition dynamic, which is absent in AR decoding where each
position is decided once and never revisited, provides a structural pathway for
small token-level preferences to compound along the denoising trajectory.

The practical relevance of this distinction is supported empirically. Activation
steering, a representation-level DLM steering baseline, fails to surpass even the
prompt-only baseline in three of four settings (Table~\ref{tab:ose_wikipol}),
whereas DLM-SWAI yields consistent gains. This suggests that operating directly
on denoising-time token distributions is an important factor behind the
effectiveness of token-level biasing in DLMs.

\section{Guideline for Human Annotators}
\begin{myrqbox}{}
One of the three sentences is always Elementary, one is Intermediate, and one is Advanced.
Even if a sentence appears somewhat corrupted or contains repetition, please ignore such issues and judge only its overall difficulty level.
Referring to the level of vocabulary used in each sentence may also be a helpful way to make your judgment.
Please enter E, I, and A for the three sentences in the order they are presented.
\end{myrqbox}

\section{Prompts for GPT Judge}
\begin{myrqbox}{OSE}
You are a strict evaluator for OneStopEnglish-style rewriting levels:
ELEMENTARY, INTERMEDIATE, ADVANCED. \\

Core principle: \\
- Judge the WRITING STYLE (simplification vs journalistic compression), not the topic. \\
- Even ELEMENTARY may contain advanced topic words; do NOT up-level based on topic vocabulary. \\

What to focus on: \\
A) Simplification signals (push toward ELEMENTARY) \\
- ``Spell-out'' paraphrases and definitions (e.g., X that does Y; ``called \dots''; explaining terms) \\
- Sentence splitting: facts spread across many short/plain sentences \\
- Basic/local cohesion: heavy reliance on and/but/so/because; list-like sequencing \\
- Repetition / low variation: repeated frames, repeated key words \\
- More explicit moral/author commentary in simple wording \\

B) Journalistic compression signals (push toward ADVANCED) \\
- Dense noun phrases and precise verbs (e.g., insists/denies/echoes/anticipates/deemed/bracing) \\
- Strong framing: setup, development, implications; effective transitions (however/nonetheless/whereas) \\
- Consistently natural collocations; little learner-like ``spell-out'' wording \\
- Information density: attribution, qualifiers, contrast, embedded clauses handled well across the text \\

Text integrity rule (important): \\
- If the prose contains obvious corruption (truncated sentences, duplicated fragments inserted mid-sentence, scrambled ordering), \\
  treat this as NOISE. Do not automatically equate noise with low level. \\
  If corruption prevents reliable judging, choose the LOWER label and lower confidence. \\

Definitions (use these exactly): \\
- ELEMENTARY: learner-simplified prose. Frequent spell-out paraphrases/definitions, short/plain sentences, basic connectors, repetition, and weaker global framing. Major grammar errors may occur but are NOT required. \\
- INTERMEDIATE: meaning is stable. Some complex sentences and connectors appear, but prose still feels simplified; fewer dense journalistic turns than ADVANCED; occasional awkwardness may remain. \\
- ADVANCED: consistently controlled, information-dense, journalistic-style prose. Compression/framing and precise word choice are sustained across the text. Minor typos are acceptable if they do not stand out. \\

Decision procedure (must follow): \\
1) Ignore non-prose (code, templates, ``Continue writing'', markdown artifacts). Judge only meaningful prose. \\
2) Choose ELEMENTARY if at least TWO simplification signals (A) are strong and persistent across the text. \\
3) Otherwise choose ADVANCED if at least THREE compression signals (B) are strong and persistent across the text. \\
4) Otherwise choose INTERMEDIATE. \\
5) Tie-break: if unsure between two labels, choose the LOWER label. \\

Output ONLY the label word: ELEMENTARY, INTERMEDIATE, or ADVANCED. Nothing else.
\end{myrqbox}

\begin{myrqbox}{WikiPol}
You are labeling the PERCEIVED POLITENESS of a request. \\

Judge how the text would sound to a typical recipient. \\
Label only the interpersonal tone of the request itself, not the writer's intent. \\
Do not infer sarcasm or hidden hostility unless it is explicit. \\

Use this rule: \\
- POLITE: clear positive politeness is explicitly present (e.g., gratitude, apology, respectful/deferential wording, clear consideration for the recipient). \\
- IMPOLITE: clear negative interpersonal force is explicitly present (e.g., pressure, entitlement, blame, complaint, accusation, adversarial or cornering wording). \\
- NEUTRAL: everything else. \\

Important: \\
- A question is not automatically POLITE. \\
- Hedging or indirectness alone is not enough for POLITE. \\
- A request without explicit courtesy should usually be NEUTRAL. \\
- Defensive self-justification or rhetorical pressure can make a request IMPOLITE. \\

Ungrammatical, awkward, or unnatural wording is NOT by itself evidence of impoliteness. \\
Judge politeness independently from fluency or grammatical quality whenever possible. \\

Output exactly one word: \\
POLITE \\
NEUTRAL \\
IMPOLITE
\end{myrqbox}

\end{document}